\documentclass{ecai}  

\usepackage{graphicx}
\usepackage{latexsym}
\usepackage{times}
\usepackage{soul}
\usepackage{url}
\usepackage[hidelinks]{hyperref}
\usepackage[utf8]{inputenc}
\usepackage[small]{caption}
\usepackage{graphicx}
\usepackage{amsmath}
\usepackage{amsthm}
\usepackage{booktabs}
\usepackage{algorithm}
\usepackage{algorithmic}
\usepackage{times}
\usepackage{soul}
\usepackage{url}
\usepackage[hidelinks]{hyperref}
\usepackage[utf8]{inputenc}
\usepackage{graphicx}
\usepackage{amsmath}
\usepackage{amssymb}
\usepackage{booktabs}
\usepackage{algorithm}
\usepackage{algorithmic}
\usepackage[switch]{lineno}
\usepackage{color}
\usepackage{array}
\usepackage{booktabs}
\usepackage{multirow}
\usepackage{bm}
\usepackage{mathrsfs}
\usepackage{enumitem}
\usepackage{amsthm}

\begin{document}

\begin{frontmatter}

\title{Improving Text Semantic Similarity Modeling through a 3D Siamese Network}

\author[A]{\fnms{Jianxiang}~\snm{Zang}\orcid{0000-0003-0236-8215}\thanks{Corresponding Author. Email: 21349110@suibe.edu.cn}}
\author[B]{\fnms{Hui}~\snm{Liu}\orcid{0000-0003-1679-6560}}

\address[A]{School of Statistics and Informatics, Shanghai University of International Business and Economics, China}

\begin{abstract}
Siamese networks have gained popularity as a method for modeling text semantic similarity. Traditional methods rely on pooling operation to compress the semantic representations from Transformer blocks in encoding, 
resulting in two-dimensional semantic vectors and the loss of hierarchical semantic information from Transformer blocks. Moreover, this limited structure of semantic vectors is akin to a flattened landscape, which restricts the methods that can be applied in downstream modeling, as they can only navigate this flat terrain.
To address this issue, we propose a novel 3D Siamese network for text semantic similarity modeling, which maps semantic information to a higher-dimensional space. The three-dimensional semantic tensors not only retains more precise spatial and feature domain information but also provides the necessary structural condition for comprehensive downstream modeling strategies to capture them.
Leveraging this structural advantage, we introduce several modules to reinforce this 3D framework, focusing on three aspects: feature extraction, attention, and feature fusion. Our extensive experiments on four text semantic similarity benchmarks demonstrate the effectiveness and efficiency of our 3D Siamese Network. 

\end{abstract}

\end{frontmatter}

\section{Introduction}~\label{introduction}
The aim of modeling text semantic similarity is to predict the degree of similarity between a pair of text sequences \cite{huang2013learning, chen2017enhanced, palangi2016deep, ruckle2020multicqa, cao2020deformer}. One of the popular methods for modeling text semantic similarity is the Siamese network, which employs dual Transformer encoders to encode two texts separately and fuse them together at the matching layer \cite{khattab2020colbert, li2021virt}. The Transformer blocks used as encoders capture hierarchical semantic information that can be viewed from two aspects: positional (spatial) domain information and feature domain information \cite{vaswani2017attention}. Shallow Transformer blocks have high global feature encapsulation, whereas deep blocks have comparatively lower global feature encapsulation. On the other hand, deep Transformer blocks have high encapsulation of local feature and positional information, while shallow blocks have low encapsulation of local feature information. 

\begin{figure}[t]
\centering
\includegraphics[width=1\linewidth]{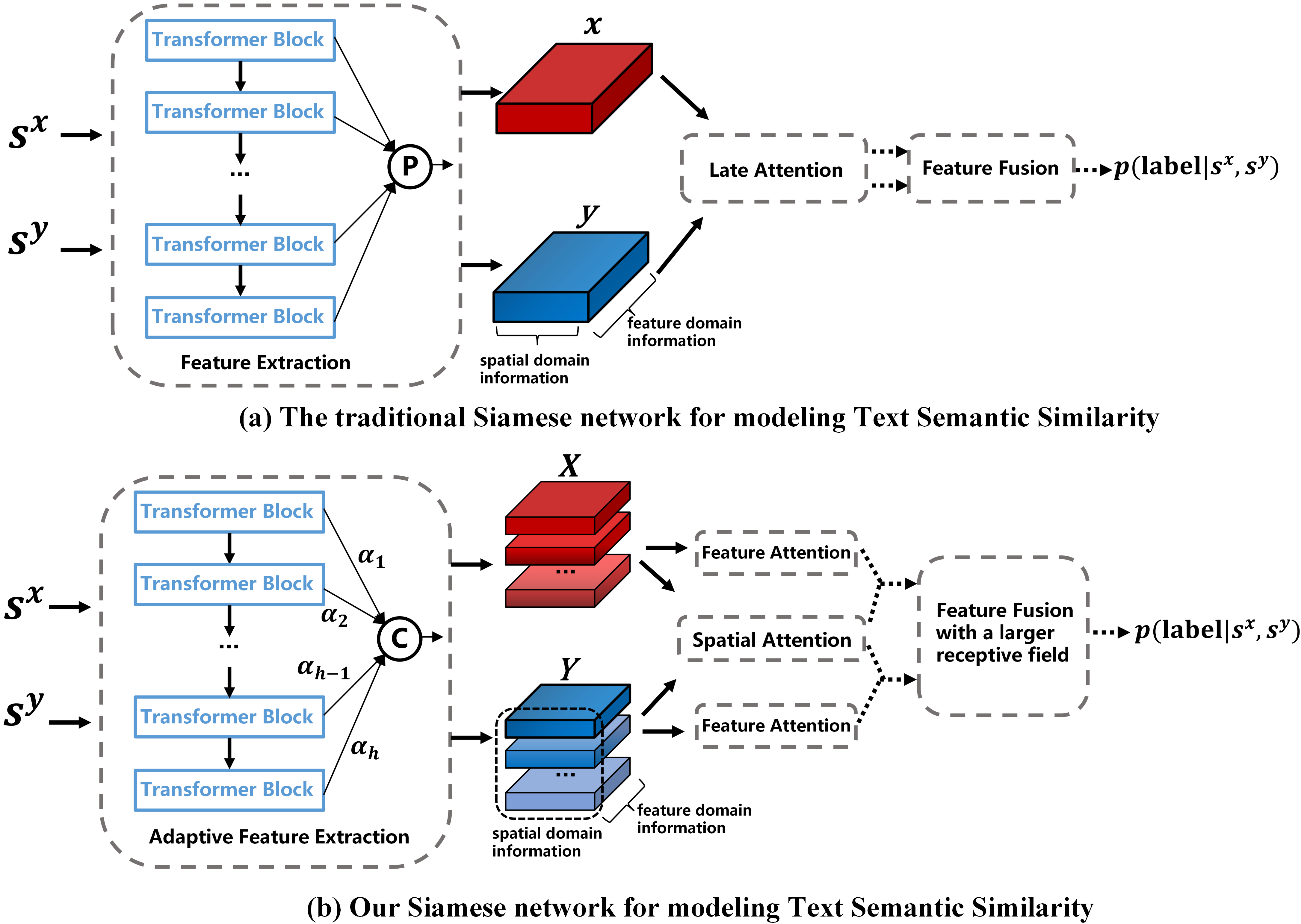}
\caption{Comparison between the traditional Siamese Network and our 3D network for modeling text semantic similarity, where 'P' denotes the pooling operation and 'C' denotes the concatenation operation.}
\label{fig1}
\end{figure}

In order to fully utilize these semantic information for modeling semantic similarity, researchers have introduced different modules into the models from three perspectives: feature extraction~\cite{reimers2019sentence,huang2017densely,gongnatural}, late attention~\cite{humeau2019poly,liu2021pay,cao2020deformer}, and feature fusion~\cite{wang2017bilateral,chen2017enhanced,reimers2019sentence,khattab2020colbert,yang2019simple,li2021virt}. However, as illustrated in Figure~\ref{fig1}(a), these methods are based on a flattened processing, pooling the representation from each Transformer, and in some cases, even solely utilizing the representation from the final Transformer block.~\cite{khattab2020colbert,humeau2019poly,ni2022sentence,li2021virt}. These processes lead to two-dimensional (sentence length, feature dimension) semantic vectors and causes a substantial loss of hierarchical semantic information from different Transformer blocks. Moreover, this limited structure of semantic vectors is akin to a flattened landscape, which restricts the methods that can be applied in downstream modeling, as they can only navigate this flat terrain.
To tackle this issue, as depicted in Figure~\ref{fig1}(b), we have implemented a novel technique that models text semantic similarity through a 3D Siamese network. This 3D network architecture maps semantic information to a higher-dimensional space, comprehensively and effectively retaining the hierarchical semantic information from Transformer blocks. For traditional methods of modeling two-dimensional semantic tensor similarity, the limitations lie in the tensor dimensions, which dictate the use of a single attention mechanism, and the reliance on global pooling during the feature fusion stage. In contrast, the resulted three-dimensional semantic tensors provides the necessary structural condition for a more diverse range of options for attention and feature fusion in downstream tasks, enabling a more comprehensive and robust capture of spatial and feature domain information.
Leveraging this structural advantage, and drawing inspiration from modules within the image processing domain~\cite{hu2018squeeze,woo2018cbam,hou2021coordinate,szegedy2016rethinking,liu2018receptive}, we suggest several modules focusing on three vital aspects to reinforce this network: feature extraction, attention, and feature fusion. Specifically, we introduce \textbf{A}daptive weighting to \textbf{E}xtract \textbf{F}eatures (\textbf{AFE}) from the representation from each transformer block, and use a stacked concatenation method to encode sentences for downstream semantic similarity modeling. The 3D Siamese network provides increased potential for attention mechanisms and feature fusion. In contrast to traditional single late attention, we simultaneously incorporate \textbf{S}patial \textbf{A}ttention (\textbf{SA}) and \textbf{F}eature \textbf{A}ttention (\textbf{FA}), forming a robust information interactor. Moreover, from a higher-dimensional standpoint, we draw upon Inception architecture and Dilated Convolution to develop a \textbf{R}eceptive \textbf{F}ield \textbf{M}odule (\textbf{RFM}) for feature fusion. 

Our main contributions resides in the following aspects:
\begin{enumerate}[label=\textbullet]
    \item We propose a 3D Siamese network that comprehensively and effectively retaining the hierarchical semantic information from Transformer blocks, and provides the necessary structural condition for comprehensive downstream semantic similarity modeling strategies to capture them.
    \item  To reinforce this 3D Siamese network, we suggest several modules for the downstream semantic similarity task that focus on three crucial aspects: feature extraction, attention, and feature fusion.
    \item Introduced modules exhibit "plug-and-play" characteristic, which contributes to the model's robust modularity and scalability. 
\end{enumerate}

\section{Related Work}~\label{related work}
Text semantic similarity modeling is a core problem in Natural Language Processing, which aims to determine the semantic relationship between two given textual inputs\cite{huang2013learning,chen2017enhanced,palangi2016deep,ruckle2020multicqa,cao2020deformer}. 

In recent years, Transformer-based pre-trained encoders have garnered significant attention for modeling text semantic similarity~\cite{vaswani2017attention}, yielding remarkable results~\cite{raffel2020exploring,devlin2019bert,xiong2020approximate}. Two primary training paradigms have emerged, based on Transformer encoders. The first entails the joint encoding of sentence pairs by Transformer modules~\cite{nogueira2019passage}, which facilitates comprehensive interaction between sentence pairs in terms of information. However, this approach results in increased computational costs and higher inference latency, posing challenges for industrial deployment. Alternatively, the second approach employs a structure with Siamese Transformer encoders~\cite{devlin2019bert,reimers2019sentence,khattab2020colbert,ni2022sentence}. This method enables offline computation of text embeddings, significantly reducing online latency~\cite{cer2018universal,reimers2019sentence}. 

To enhance the performance of Siamese networks, researchers have proposed corresponding improvements in three areas: feature extraction, late attention, feature fusion. In terms of feature extraction, researchers~\cite{reimers2019sentence} compared the effects of three pooling fusion methods on the performance of Siamese BERT. The average-pooling strategy can extract more abundant information compared with max pooling. In addition to pooling operation, DenseNet~\cite{huang2017densely,gongnatural} has also been used for feature extraction~\cite{cer2018universal}, largely preserving the information of the original text features. In order to enhance the model's interaction capability, researchers have introduced numerous late attention strategies, such as the cross-attention layer~\cite{humeau2019poly}, the MLP layer~\cite{liu2021pay}, and the Transformer layer~\cite{cao2020deformer}. Regarding feature fusion, tensor cross-fusion~\cite{wang2017bilateral,chen2017enhanced,reimers2019sentence} and pooling fusion~\cite{chen2017enhanced,reimers2019sentence,khattab2020colbert} are the most commonly used methods for feature fusion. Based on these,~\cite{yang2019simple} propose an improved mechanism for information fusion and interaction that compares local and aligned representations from three perspectives. ~\cite{li2021virt} propose VIRT-Adapted Interaction which performs feature extraction and information exchange simultaneously.

As mentioned in Section~\ref{introduction}, these methods are based on a uniform and flattened processing, pooling the representation from each Transformer block during encoding,~\cite{reimers2019sentence,khattab2020colbert,humeau2019poly,ni2022sentence,li2021virt}, the limitation of which lies in the insufficient extraction of these two types of information, as well as the constraints imposed on the downstream choices of similarity modeling. To address this issue, we have adopted a novel approach modeling text semantic similarity through a three-dimensional Siamese network and suggest several modules to reinforce it.

\section{Methodology}

The task of modeling text semantic similarity can be described as following. Given the input sentence pair $\bm{s}^x$ and $\bm{s}^y$, the training goal of the model is to train a classifier $\xi$ that computes conditional probabilities $P({\rm label}|\bm{s}^x,\bm{s}^y)$ to predict the relationship between the output sentence pairs based on the output probabilities. The ${\rm label}\in \Omega$ represents different degrees of semantic similarity.

\begin{equation}
    P({\rm label}|\bm{s}^x,\bm{s}^y)=\xi(\bm{s}^x,\bm{s}^y)
\end{equation}

In this paper, we introduce a novel 3D Siamese network for text semantic similarity modeling, emphasizing feature extraction, attention, and feature fusion. The 3D Siamese network consists of the following modules we introduced: Adaptive Feature Extraction (AFE), Spatial Attention \& Feature Attention (SA\&FA), and Receptive Field Module (RFM). AFE comprehensivly and efficiently preserves semantic information from different Transformer blocks. SA captures long-range dependencies between sentence pairs, while FA dynamically adjusts feature weights within certain sentence for better discrimination. We combine SA and FA for robust information interactor. Finally, RFM leverages the Inception architecture and Dilated Convolution layers to capture a larger receptive field.

\subsection{Adaptive Feature Extraction}	
 
Compared to the traditional method of using pooling operations to compress semantic representations from transformer blocks,, we propose the use of trainable Adaptive weights for Feature Extraction (AFE) in each block. These weighted representations are concatenated in a three-dimensional form, delivering the most comprehensive information for downstream task modeling.

Given the input sentence pair $\bm{s}^x=[\bm{s}^x_1,\bm{s}^x_2...\bm{s}^x_{l_x}]$ and $\bm{s}^y=[\bm{s}^y_1,\bm{s}^y_2...\bm{s}^y_{l_y}]$, we use Transformer blocks to acquire representations of the input sentences. Assuming there are $H$ Transformer blocks, the semantic representations of the sentence pair ${ \bm{x}^h_i}$ and ${\bm{y}^h_j}$ can be obtained in the $h^{th}$ Transformer block.

\begin{equation}
\begin{aligned}
    \bm{x}^h_i&={\rm TransformerBlock}^h(\bm{s}^x,i),i \in [1,2...l_x]\\
    \bm{y}^h_j&={\rm TransformerBlock}^h(\bm{s}^y,j),j \in [1,2...l_y]
\end{aligned}
\end{equation}

Our goal is to obtain a semantic tensor comprising the representations from each Transformer block, concatenated using trainable weights. We define our unnormalized attention vectors $\widetilde{\bm{x}}^h_i$ and $\widetilde{\bm{y}}^h_j$ as follows:

\begin{equation}
\begin{aligned}
\widetilde{\bm{x}}^h_i&=\sigma(W^h_2({\rm ReLU}(W^h_1\bm{x}^h_i+b^h_1))+b^h_2)\\    \widetilde{\bm{y}}^h_j&=\sigma(W^h_2({\rm ReLU}(W^h_1\bm{y}^h_j+b^h_1))+b^h_2)    
\end{aligned}   
\end{equation}

where $W^h_1$, $W^h_2$, $b^h_1$, $b^h_2$ are trainable parameters for the representation from the $h^{th}$ Transformer block. $\sigma$ refers to the sigmoid activation function. $\rm ReLU$ refers to the ReLU activation function. We calculate the normalized attention weights $\bm{\alpha}^h_i$ and $\bm{\beta}^h_j$:

\begin{equation}
\begin{gathered}
     \bm{\alpha}^h_i=\frac{\widetilde{\bm{x}}^h_i}{\sum_{i=1}^{l_x}\widetilde{\bm{x}}^h_i},\quad
    \bm{\beta}^h_j=\frac{\widetilde{\bm{y}}^h_j}{\sum_{j=1}^{l_y}\widetilde{\bm{y}}^h_j}   
\end{gathered}
\end{equation}

These weights are applied to perform a weighted concatenation on the representations from each transformer block, producing final semantic tensors $\{\bm{X},\bm{Y}\}$ that are utilized for further modeling of semantic similarity in downstream tasks.

\begin{equation}
\begin{aligned}
      \bm{X}=[\bm{\alpha}^1_i\bm{x}^1_i;\bm{\alpha}^2_i\bm{x}^2_i...;\bm{\alpha}^H_i\bm{x}^H_i]\\
    \bm{Y}=[\bm{\beta}^1_j\bm{y}^1_j;\bm{\beta}^2_j\bm{y}^2_j;...;\bm{\beta}^H_j\bm{y}^H_j]  
\end{aligned}
\end{equation}

In above equation both $\bm{X}$ and $\bm{Y}\in \mathbb{R}^{H\times L\times D}$. $D$ is the dimension of the sentence vector. $H$ is the number of Transformer blocks, and $L$ is the length of the sentence vector. $[;]$ is the concatenation operation.

\subsection{Spatial Attention \& Feature Attention}	
As we have mentioned, traditional methods of modeling text semantic similarity employ consistent pooling compression of the representations generated by transformer blocks. As a result, they can only adopt a single late attention to model the spatial and feature domain information between sentences. In contrast, our approach takes a 3D perspective and allows for the separate modeling of Spatial Attention (SA) and Feature Attention (FA). By fusing the semantic tensors that have been enhanced through these attention mechanisms for feature fusion, our approach forms a more powerful information interactor. 
Specifically, Spatial Attention effectively learns the inter-dependencies between various positions of two semantic tensors, enhancing its ability to capture long-range dependencies between sentence pairs. Meanwhile, Feature Attention is capable of discerning dependencies among features within the semantic tensor, dynamically adjusting the weights for each feature and thereby extracting more discriminative features. For the semantic tensors $\{\bm{X}$,$\bm{Y}\}\in \mathbb{R}^{H\times L \times D}$, we fuse the results of two attention-enhanced processes using either feature-level or element-wise multiplication to obtain the semantic tensors $\{\bm{X'}$,$\bm{Y'}\}\in \mathbb{R}^{H\times L \times D'}$.

\begin{equation}
\begin{aligned}
     \bm{X}'&={\rm SA}(\bm{X},\bm{Y})\odot {\rm FA}(\bm{X})\\
    \bm{Y}'&={\rm SA}(\bm{Y},\bm{X})\odot {\rm FA}(\bm{Y})\label{eq.sa&fa}   
\end{aligned}
\end{equation}

where SA and FA refer to Spatial Attention Module and Feature Attention Module respectively. $\odot$ refers to elements-wise multiplication.

\subsubsection{Spatial Attention}

\begin{figure}[t]
\centering
\includegraphics[width=1\linewidth]{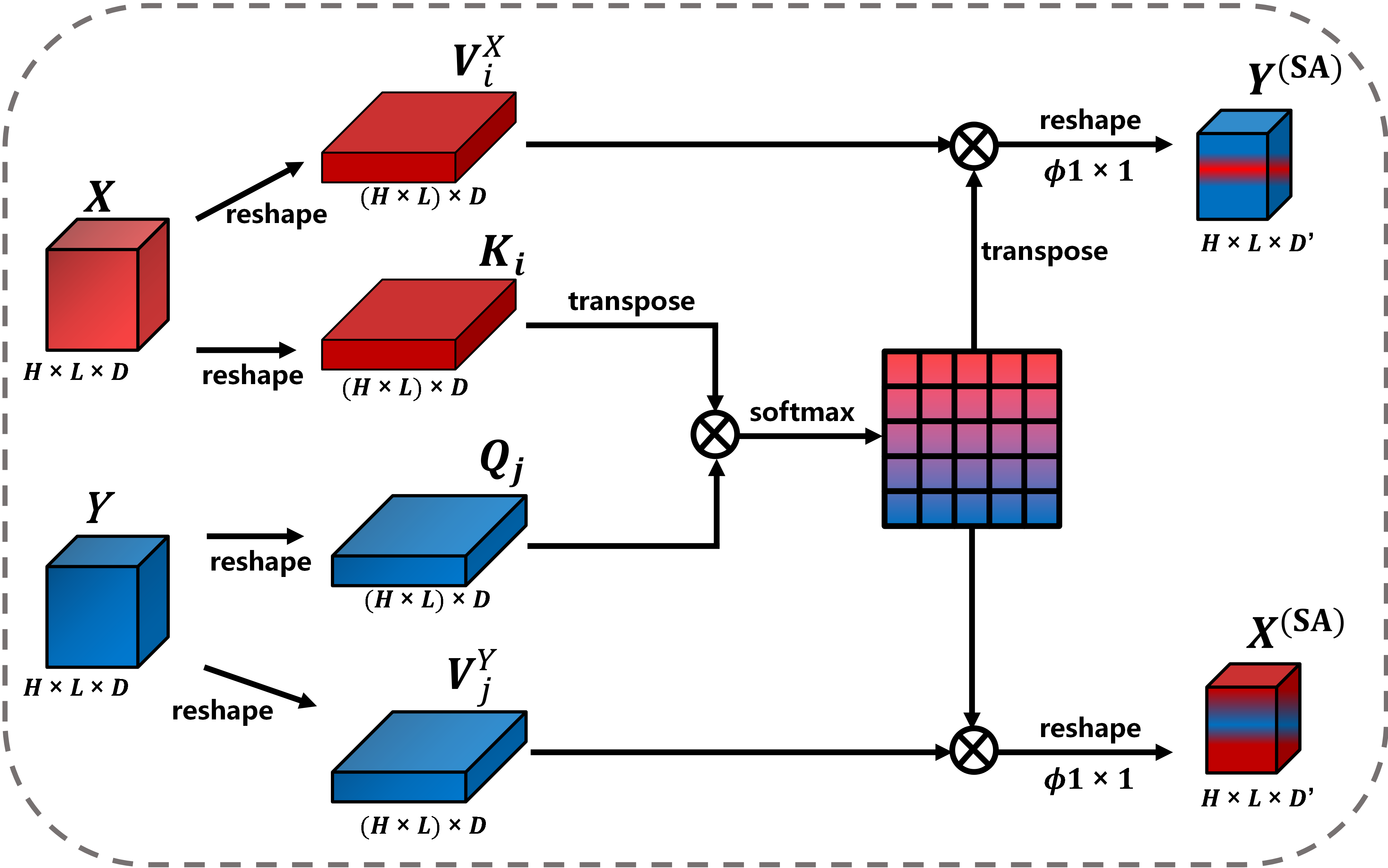}
\caption{Overview of Spatial Attention.}
\label{fig.sa}
\end{figure}

Spatial Attention enables the modeling of inter-dependencies between different positions of semantic tensor pairs, which enhances the ability of our model to capture long-range dependencies between pairs of sentences. Let $\{\bm{X},\bm{Y}\}\in \mathbb{R}^{H\times L\times D}$ denote the semantic tensors that have been processed by the Adaptive Feature Fusion and Extraction module. We first reshape them to $\bm{K}_i$ and $\bm{Q}_j$ with the shape of $N \times D$, where $N = H \times L$. Different from self-attention, our query matrix $\bm{Q}$ and key matrix $\bm{K}$ interact with each other for cross-sentence information communication. Therefore, as illustrated in Figure~\ref{fig.sa} , we design two branches to process sentence pair's representations respectively. After that, we perform a matrix multiplication between the transpose of $\bm{K}$ and $\bm{Q}$  and transpose this calculation result to obtain the semantic tensor of another branch. Finally, we  apply a softmax layer on them to calculate $\bm{M}_{ji}^Y$ and $\bm{M}_{ij}^X$ respectively.

\begin{equation}
\begin{gathered}
\bm{M}^Y_{ji}=\frac{{\rm exp}(\bm{K}_i^T\cdot \bm{Q}_{j})}{\sum_{k=1}^{N}{{\rm exp}(\bm{K}_i^T\cdot \bm{Q}_{k})}},\quad
\bm{M}^X_{ij}=\frac{{\rm exp}(\bm{K}_i^T\cdot \bm{Q}_{j})}{\sum_{k=1}^{N}{{\rm exp}(\bm{K}_k^T\cdot \bm{Q}_{j})}} \label{eq.kq}
\end{gathered}
\end{equation}

Here, similarity matrix $\bm{M}^Y_{ji}$ measures the $\bm{Y}$'s $j^{th}$ position’s impact on the $i^{th}$ position of $\bm{X}$. The more similar the feature representations of the two positions are, the larger the correlation between them becomes. The function of $\bm{M}^X_{ij}$ is similar with Equation~\ref{eq.kq}.

Meanwhile, we reshape $\bm{X}$ and $\bm{Y}$ reshape them to $\bm{V}^X_i$ and $\bm{V}^Y_j$ with the shape of $N \times D$, where $N = H \times L$. Then we perform a matrix multiplication between $\bm{V}_j^Y$ and $\bm{M}_{ji}^Y$, after reshaping them to the shape of $N\times D$. For another branch, similar to the above calculation process, we perform a matrix multiplication between $\bm{V}_i^X$ and $\bm{M}_{ij}^X$. 

\begin{equation}
\begin{aligned}
  \bm{X}^{(\rm SA)}_i&=\sum_{j=1}^{N}{\bm{M}^Y_{ji}\cdot \bm{V}^Y_j},\quad {i \in [1,...N]}  \\ \bm{Y}^{(\rm SA)}_j&=\sum_{i=1}^{N}{\bm{M}^X_{ij}\cdot \bm{V}^X_i},\quad {j \in [1,...N]}\label{eq.sam}
\end{aligned}
\end{equation}

Where $\bm{X}^{(\rm SA)}_i$ is the weighted sum of $\{\bm{Y}_j\}_{j=1}^{N}$ with $\bm{M}^Y_{ji}$. Intuitively, the purpose of Spatial Attention is to use the elements in $\{\bm{Y}_j\}_{j=1}^{N}$ that are related to $\bm{X}^{(\rm SA)}_i$ to represent $\bm{X}^{(\rm SA)}_i$. The same is performed for another branch with Equation~\ref{eq.sam}. 
Finally, we apply reshape operations to the results and then leverage a $1 \times 1$ convolution layer to get the outputs $\bm{X}^{(\rm SA)}$ and $\bm{Y}^{(\rm SA)}$, where $\bm{X}^{(\rm SA)},\bm{Y}^{(\rm SA)} \in \mathbb{R}^{H\times L\times D'}$ and $D'$ is the feature dimension of low dimensional mapping space.

\subsubsection{Feature Attention}

\begin{figure}[t]
\centering
\includegraphics[width=1\linewidth]{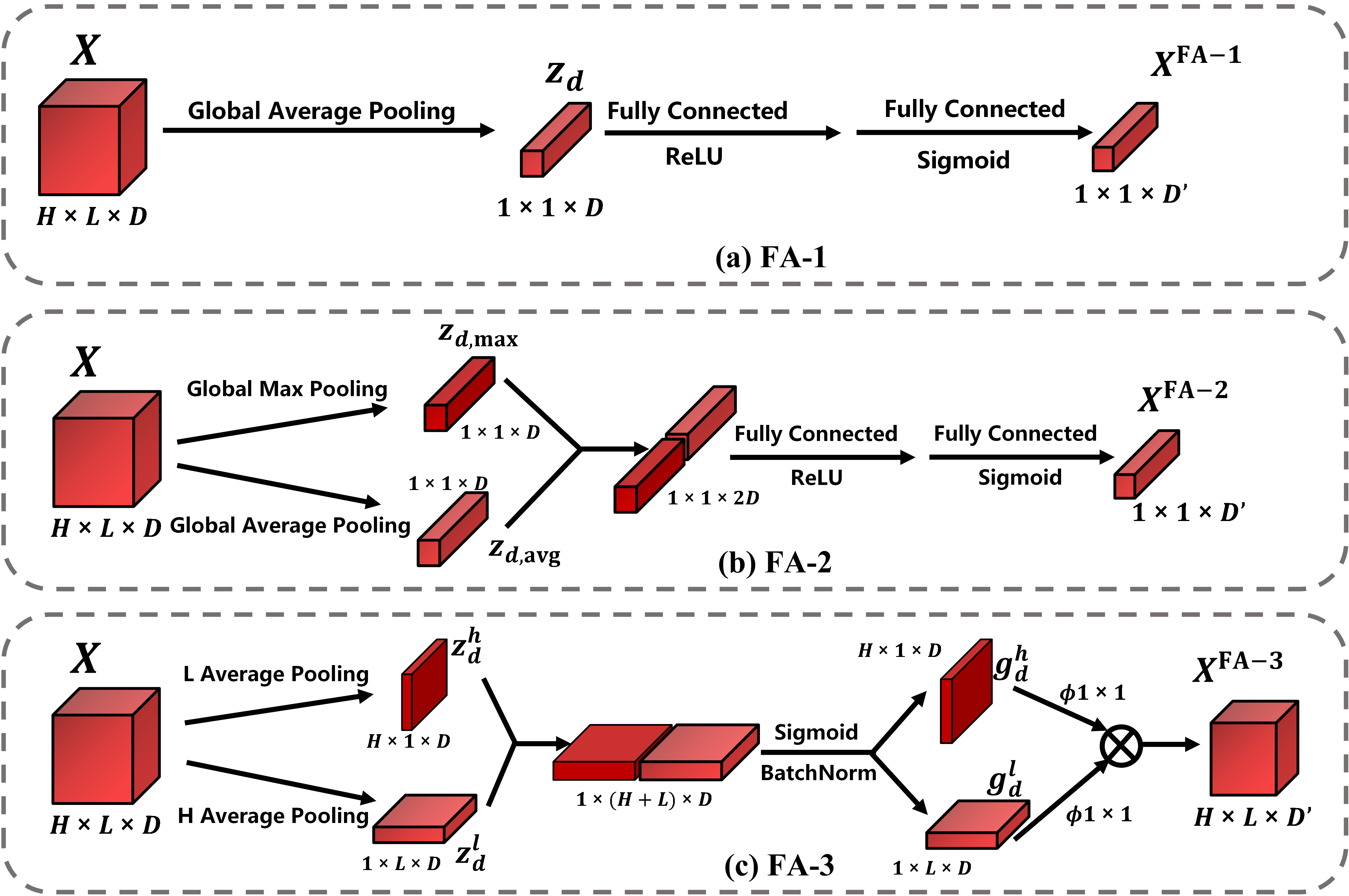}
\caption{Overview of three forms of Feature Attention.}
\label{fig.fa}
\end{figure}

The Feature Attention is a computational unit designed to learn the inter-dependencies between features. It takes a semantic tensor $\bm{X} = [\bm{x}_1, \bm{x}_2,..., \bm{x}_D ] \in \mathbb{R}^{H\times L\times D}$ or $\bm{Y} = [\bm{y}_1, \bm{y}_2,..., \bm{y}_D ] \in \mathbb{R}^{H\times L\times D}$ as input and produces transformed tensor $\bm{X}^{(\rm FA)}$ or $\bm{Y}^{(\rm FA)} \in \mathbb{R}^{H\times L\times D'}$  with augmented representations of the same size as $\bm{X}^{(\rm SA)}$ or $\bm{Y}^{(\rm SA)}$. Inspired by the Squeeze-and-Excitation Networks~\cite{hu2018squeeze,woo2018cbam,hou2021coordinate}, Feature Attention can be modeled in two stages, the squeeze stage and the excitation stage. Specifically, the Squeeze stage is responsible for compressing the input semantic tensor to obtain a spatial context descriptor~\cite{heikkila2009description}, which is a process of extracting global information. The Excitation performs non-linear transformations on the compressed semantic tensor to obtain a weight vector for each feature. We propose three forms of FA. To simplify illustration, we will only show the processing of FA in the branch of input $\bm{X}$ as example in Figure~\ref{fig.fa}.
\begin{enumerate}
    \item 
    For FA-1, given the input $\bm{X}$, to open up the correlation between feature and spatial domain information, we squeeze this semantic tensor using average pooling. The squeeze step for the $d^{th}$ feature can be formulated in Equation~\ref{eq.squezze1}
    
    \begin{equation}
    \bm{z}_d=\frac{1}{H\times L}\sum_{i=1}^H\sum_{j=1}^L\bm{x}_d(i,j)\label{eq.squezze1}
    \end{equation}

    where $\bm{z}_d$ is the spatial context descriptor associated with the $d^{th}$ feature. This squeeze operation makes collecting global information possible. 

     Next, we utilize the feedforward layers to obtain the semantic tensor $\bm{X}^{(\rm FA-1)} \in \mathbb{R}^{H\times L \times D'}$, fully capturing feature-wise dependencies.

    \begin{equation}
    \bm{X}^{(\rm FA-1)}=\sigma(W_2({\rm ReLU}(W_1\bm{z}+b_1))+b_2)\label{eq.extraction}
    \end{equation}

    where $W_1$, $W_2$, $b_1$, $b_2$ are trainable parameters, $\sigma$ is the sigmoid function. 
    \item
    For FA-2, compared to FA-1, we argue that utilizing max pooling captures another crucial aspect of object characteristics, which enables us to infer Feature Attention with even greater precision. Firstly, we utilize average pooling and max pooling operations to generate two different spatial context descriptors: $\bm{z}_{d,{\rm avg}}$ and $\bm{z}_{d,{\rm max}}$.
    \begin{equation}
    \bm{z}_{d,{\rm avg}}=\frac{1}{H\times L}\sum_{i=1}^H\sum_{j=1}^L\bm{x}_d(i,j)\label{eq.squeeze2 avg}
    \end{equation}    

    \begin{equation}
    \bm{z}_{d,{\rm max}}={\rm max}_{i=1}^H{\rm max}_{j=1}^L\bm{x}_d(i,j)\label{eq.squeeze2 max}
    \end{equation}

    Similar to FA-1,  Both descriptors are concatenated then forwarded to the feedforward layers to produce the final semantic tensor $\bm{X}^{(\rm FA-2)} \in \mathbb{R}^{H\times L \times D'}$.

    \begin{equation}
    \bm{X}^{(\rm FA-2)}=\sigma(W_2({\rm ReLU}(W_1([\bm{z}_{d,{\rm avg}};\bm{z}_{d,{\rm max}}])+b_1))+b_2)\label{eq.extraction2}
    \end{equation} 
    
    \item
    For FA-3, we further enhance the process by accurately model the relationships between features located at longer distances in a spatial context while also taking into account their specific positions within that space. To achieve this, we factorize the global pooling operation formulated in the original method into a pair of 1D feature encoding operations. Specifically, given an input $\bm{X}$, we use two pooling kernels with different spatial extents, $(H, 1)$ and $(1, L)$, to encode each feature along the horizontal and vertical coordinates, respectively. This allows us to compute the output of the $d^{th}$ feature at $h^{th}$ Transformer block  and the output of the $d^{th}$ feature at the $l^{th}$ position in the sentence, formulated in Equation~\ref{eq.squeeze3}.
    
    \begin{equation}
    \begin{gathered}
    \bm{z}_{d}^h(h)=\frac{1}{L}\sum_{i=0}^L\bm{x}_d(i,j), \quad
    \bm{z}_{d}^l(l)=\frac{1}{H}\sum_{j=0}^H\bm{x}_d(i,j)\label{eq.squeeze3}
    \end{gathered}
    \end{equation} 

    The above two transformations enable a global receptive field and encode precise positional information. Compared to FA-1 and FA-2, we not only want to make full use of the captured spatial information but also to effectively capture inter-feature relationships. Specifically, given the descriptors produced by Equation~\ref{eq.squeeze3}, we apply sigmoid activation and batch normalization processing to the concatenated result.

    \begin{equation}
    \bm{g}={\rm BatchNorm}({\sigma}([\bm{z}_{d}^h;\bm{z}_{d}^l]))
    \end{equation}

    where $\bm{g} \in \mathbb{R}^{1 \times (H+L)\times D}$ is the intermediate descriptor that encodes spatial information in both the horizontal direction and the vertical direction. Here, $D'$ is the feature dimension of low dimensional mapping. We then split $\bm{g}$ along the spatial dimension into two separate tensors $\bm{g}^h \in \mathbb{R}^{H\times 1 \times D}$ and $\bm{g}^l \in \mathbb{R}^{1 \times L\times D}$ . Another two $1\times 1$ convolution layers $\phi_h$ , $\phi_l$ are utilized to reshape $\bm{g}^h$ and $\bm{g}^l$ to tensors with the same feature dimension to $D'$. Finally we leverage the matrix multiplication between these two tensor, and obtain $\bm{X}^{(\rm FA-3)} \in \mathbb{R}^{H\times L \times D'}$. 

    \begin{equation}
    \bm{X}^{\rm (FA-3)}=\phi_h(\bm{g}^h)\cdot \phi_l(\bm{g}^l)
    \end{equation}
Where $\cdot$ denotes the matrix multiplication.

\end{enumerate}

\begin{figure}[t]
\centering
\includegraphics[width=1\linewidth]{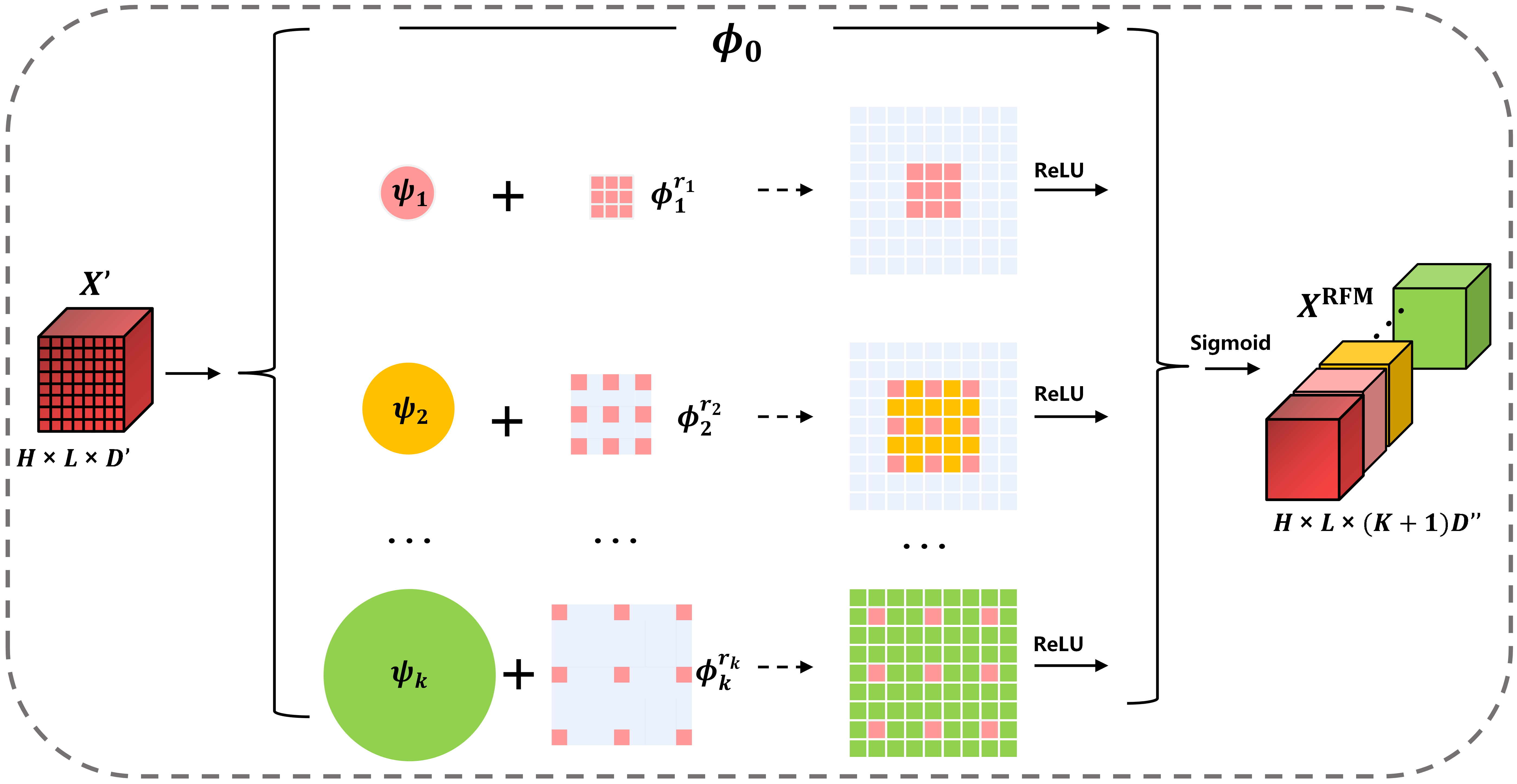}
\caption{Overview of Receptive Field Module}
\label{fig.rfm}
\end{figure}

\begin{table*}
\begin{tabular}{ccccccccccc}
\hline
\multicolumn{1}{c|}{\multirow{2}{*}{Methods}}    & \multicolumn{3}{c|}{Modules}                                                                                     & \multicolumn{5}{c|}{Accuracy(Test)}                                                                                                            & \multicolumn{2}{c}{Latency\&Params}                       \\ \cline{2-11} 
\multicolumn{1}{c|}{}                            & \multicolumn{1}{c|}{Feature Extraction} & \multicolumn{1}{c|}{Attention} & \multicolumn{1}{c|}{Feature Fusion}   & \multicolumn{1}{c|}{QQP}   & \multicolumn{1}{c|}{MRPC}  & \multicolumn{1}{c|}{SNLI}  & \multicolumn{1}{c|}{MNLI}  & \multicolumn{1}{c|}{Avg}   & \multicolumn{1}{c|}{Latency} & \multicolumn{1}{c}{Params} \\ \hline
\multicolumn{1}{c|}{SBERT~\cite{reimers2019sentence}}                & Pooling            & -       & \multicolumn{1}{c|}{Max\&Avg Pooling} & 80.81 & 70.23 & 82.88 & 72.35 & \multicolumn{1}{c|}{76.57} & 0.2ms   & \multicolumn{1}{c}{6.1M}  \\ 
\multicolumn{1}{c|}{ColBERT~\cite{khattab2020colbert}}                     & Pooling            & LA       & \multicolumn{1}{c|}{Max\&Avg Pooling} & 86.31 & 74.23 & 86.12 & 76.23 & \multicolumn{1}{c|}{81.22} & 0.3ms   & \multicolumn{1}{c}{6.1M}  \\ \hline
\multicolumn{1}{c|}{\multirow{8}{*}{Our Networks}} & AFE                                     & SA                             & \multicolumn{1}{c|}{Max\&Avg Pooling}                      & 89.34                      & 77.03                      & 88.43                      & 77.13                      & \multicolumn{1}{c|}{83.23}                      & 0.5ms                        & 6.1M                       \\
\multicolumn{1}{c|}{}                            & AFE                                     & SA,FA-1                        & \multicolumn{1}{c|}{Max\&Avg Pooling}                      & 89.45                      & 77.04                      & 89.10                       & 77.11                      & \multicolumn{1}{c|}{84.06}                      & 0.5ms                        & 6.1M                       \\
\multicolumn{1}{c|}{}                            & AFE                                     & SA,FA-2                        & \multicolumn{1}{c|}{Max\&Avg Pooling}                      & 89.48                      & 77.02                      & 89.14                      & 77.13                      & \multicolumn{1}{c|}{83.44}                      & 0.5ms                        & 6.1M                       \\
\multicolumn{1}{c|}{}                            & AFE                                     & SA,FA-3                        & \multicolumn{1}{c|}{Max\&Avg Pooling}                      & 89.46                      & 77.04                      & 89.02                      & 77.21                      & \multicolumn{1}{c|}{83.43}                      & 0.5ms                        & 6.1M                       \\
\multicolumn{1}{c|}{}                            & AFE                                     & SA                             & \multicolumn{1}{c|}{RFM}                                   & 90.02                      & 78.10                      & 89.13                      & 77.24                      & \multicolumn{1}{c|}{83.62}                      & 0.6ms                        & 6.4M                       \\
\multicolumn{1}{c|}{}                            & AFE                                     & SA,FA-1                        & \multicolumn{1}{c|}{RFM}                                   & 90.09                      & 78.11                      & 89.17                      & 77.24                      & \multicolumn{1}{c|}{83.65}                      & 0.6ms                        & 6.4M                       \\
\multicolumn{1}{c|}{}                            & AFE                                     & SA,FA-2                        & \multicolumn{1}{c|}{RFM}                                   & 90.11                      & 78.05                      & 89.11                      & 77.23                      & \multicolumn{1}{c|}{83.63}                      & 0.6ms                        & 6.4M                       \\
\multicolumn{1}{c|}{}                            & AFE                                     & SA,FA-3                        & \multicolumn{1}{c|}{RFM}                                   & 90.23                      & 78.18                      & 89.41                      & 77.28                      & \multicolumn{1}{c|}{83.78}                      & 0.6ms                        & 6.4M                       \\ \hline
\multicolumn{4}{c|}{BERT(interaction-based)~\cite{vaswani2017attention}            }                                                                                                                         & 91.51 & 87.81 & 90.01 & 86.50 & \multicolumn{1}{c|}{88.96} & 27.4ms  & \multicolumn{1}{c}{106M}   \\ \hline 
\end{tabular}
\caption{Performance comparison on four benchmarks.}
    \label{tab.main}
\end{table*}

\subsection{Feature Fusion with a Larger Receptive Field}

From our 3D perspective, we leverage a Receptive Field Module (RFM) to perform feature fusion. Inspired by the receptive field of the human vision~\cite{liu2018receptive}~\footnote{The style of the illustrations follows the approach in reference~\cite{liu2018receptive}.}, we adopt the Inception architecture~\cite{szegedy2016rethinking,szegedy2015going,liu2018receptive}, which utilizes a multi-branch structure consisting of convolutional layers with different kernel sizes. In addition, we use Dilated Convolutional layers~\cite{chen2018encoder,chen2017deeplab,gu2020context}, which have previously been employed in the segmentation algorithm Deeplab, to further expand the receptive field.

Figure~\ref{fig.rfm} illustrates the multi-branch structure of RFM, where only the modeling of a single semantic tensor $\bm{X'}$ is presented in the figure. Specifically, RFM takes the input semantic tensors ${\bm{X}',\bm{Y}'}\in \mathbb{R}^{H\times L \times D'}$ and constructs receptive field blocks by using multiple convolution kernels $\psi$ of different sizes. This enables the network to capture feature information at different scales. Next, we perform Dilated Convolutions $\phi^r$ on $\psi(\bm{X}')$ and $\psi(\bm{Y}')$, which output multiple semantic tensors with different dilation rates. 
Assuming there are $k$ different dilation rates for the convolution with the same shape, $k$ corresponding semantic tensors can be obtained. In addition to Dilated Convolutions, a $1\times 1$ convolutional layer can also be used to encode the feature vectors, resulting in a semantic tensor with global contextual information. This feature vector can help the model understand the global semantic information. Finally, all output feature vectors are concatenated to obtain the final results $\bm{X}^{\rm RFM}$ and $\bm{Y}^{\rm RFM} \in \mathbb{R}^{H\times L\times (k+1)D''}$, which can be formulated in Equation~\ref{eq.rfm1} and Equation~\ref{eq.rfm2}.

\begin{equation}
\begin{aligned}
    \bm{X}^{\rm RFM}_i&={\rm ReLU}(\phi_i^{r_i}(\psi_{i}(\bm{X}'))),i \in [1,2...,k]\\
    \bm{Y}^{\rm RFM}_j&={\rm ReLU}(\phi_j^{r_j}(\psi_{j}(\bm{Y}'))),j \in [1,2...,k] \label{eq.rfm1}       
\end{aligned} 
\end{equation}

\begin{equation}
\begin{aligned}      
    \bm{X}^{\rm RFM}&={\sigma}([\bm{X}^{\rm RFM}_1;...;\bm{X}^{\rm RFM}_k;\phi_0(\bm{X}')]) \\
    \bm{Y}^{\rm RFM}&={\sigma}([\bm{Y}^{\rm RFM}_1;...;\bm{Y}^{\rm RFM}_k;\phi_0(\bm{Y}')])\label{eq.rfm2}    
\end{aligned} 
\end{equation}

where $\psi$ denotes the $i^{th}$ convolution with different kernel size. $\phi_i^{r_i}$ denotes the $i^{th}$ convolution with $r_i$ dilation rates, $\phi_0$ is the $1\times 1$ convolution layer. The kernels among $\phi_i^{r_i},i \in [1,2...k]$ have the same size while the kernels among $\psi_i,i \in [1,2...k]$ have different sizes to capture complex features. The specific process of $\phi^r_{A \rightarrow B}$ is shown in Equation~\ref{eq.dc}. The convolution kernel $K \in \mathbb{R}^{k_h \times k_l \times D \times D'}$ is a four-dimensional tensor, where $k_h$ and $k_l$ denote the size of the convolution kernel, and $D$ and $D'$ denote the size of input features and output features, respectively.

\begin{equation}
B(i, j, d') = \sum_{m=0}^{k_h-1} \sum_{n=0}^{k_l-1} \sum_{d=0}^{d-1} A(i + m \cdot r, j + n \cdot r, d) \cdot K(m, n, d, d')~\label{eq.dc}
\end{equation}

Finally, we concatenate the two semantic tensors along with their element-wise multiplication results, and input them into a global average pooling layer and a fully connected layer to compute semantic similarity, where $\rm {GAP}(\cdot)$ denotes the global average pooling operation.

\begin{equation}
\begin{aligned}
    \bm{v}={\rm GAP}([\bm{X}^{\rm RFM};\bm{Y}^{\rm RFM};\bm{X}^{\rm RFM}\odot \bm{Y}^{\rm RFM}])  \\ 
    P({\rm label}|\bm{s}^x,\bm{s}^y)={\rm softmax}(W_2({\rm ReLU}(W_1(\bm{v})+b_1))+b_2)
\end{aligned}    
\end{equation}

\section{Experimental Setup}
\subsection{Datasets}
We adopt the following four benchmarks for evaluation.

\textbf{QQP (Quora Question Pairs)}~\cite{iyer2017first}:
QQP is a dataset of around 400,000 question pairs from Quora. The goal is to determine if the questions have the same intent, assessing performance on semantic similarity and natural language understanding.

\textbf{MRPC (Microsoft Research Paraphrase Corpus)}~\cite{dolan2005automatically}:
MRPC has 5,800 English sentence pairs from online news. The task is to identify if the pairs are paraphrases, testing textual entailment and semantic similarity.

\textbf{SNLI (Stanford Natural Language Inference)}~\cite{bowman2015large}:
SNLI is a large textual entailment dataset with 570,000 English sentence pairs and labels for entailment relationships (entailment, contradiction, neutral). It evaluates natural language inference, determining if a hypothesis holds based on a premise.

\textbf{MNLI (Multi-Genre Natural Language Inference))}~\cite{williams2018broad}:
MNLI extends SNLI with 430,000 sentence pairs from various domains and genres. Like SNLI, it assesses natural language inference but covers a broader range of tasks.

\subsection{Baselines}
We utilize three representative semantic similarity models as baselines.

\textbf{SBERT}~\cite{reimers2019sentence}:
 Siamese BERT is a Siamese architecture that uses pre-trained BERT to separately produce embeddings of two inputs. The output embeddings of two sequences are
concatenated to give final predictions.

\textbf{ColBERT}~\cite{khattab2020colbert}:
 ColBERT serves as an efficient model for semantic similarity tasks, utilizing its late interaction mechanism to process queries and documents independently, enabling scalability. By employing dense vector representations, it effectively captures richer semantic relationships between text pairs.

\textbf{BERT(interaction-based)}~\cite{vaswani2017attention}:
 As mentioned in Section 2, BERT(interaction-based) is an interaction-based model rather than utilizing a Siamese framework. Its underlying principle involves concatenating sentence pairs and encoding them through BERT for subsequent similarity prediction. While this approach enables rich interactions, it also introduces a significant number of training parameters and inference latency.

\begin{figure}[t]
\centering
\includegraphics[width=1\linewidth]{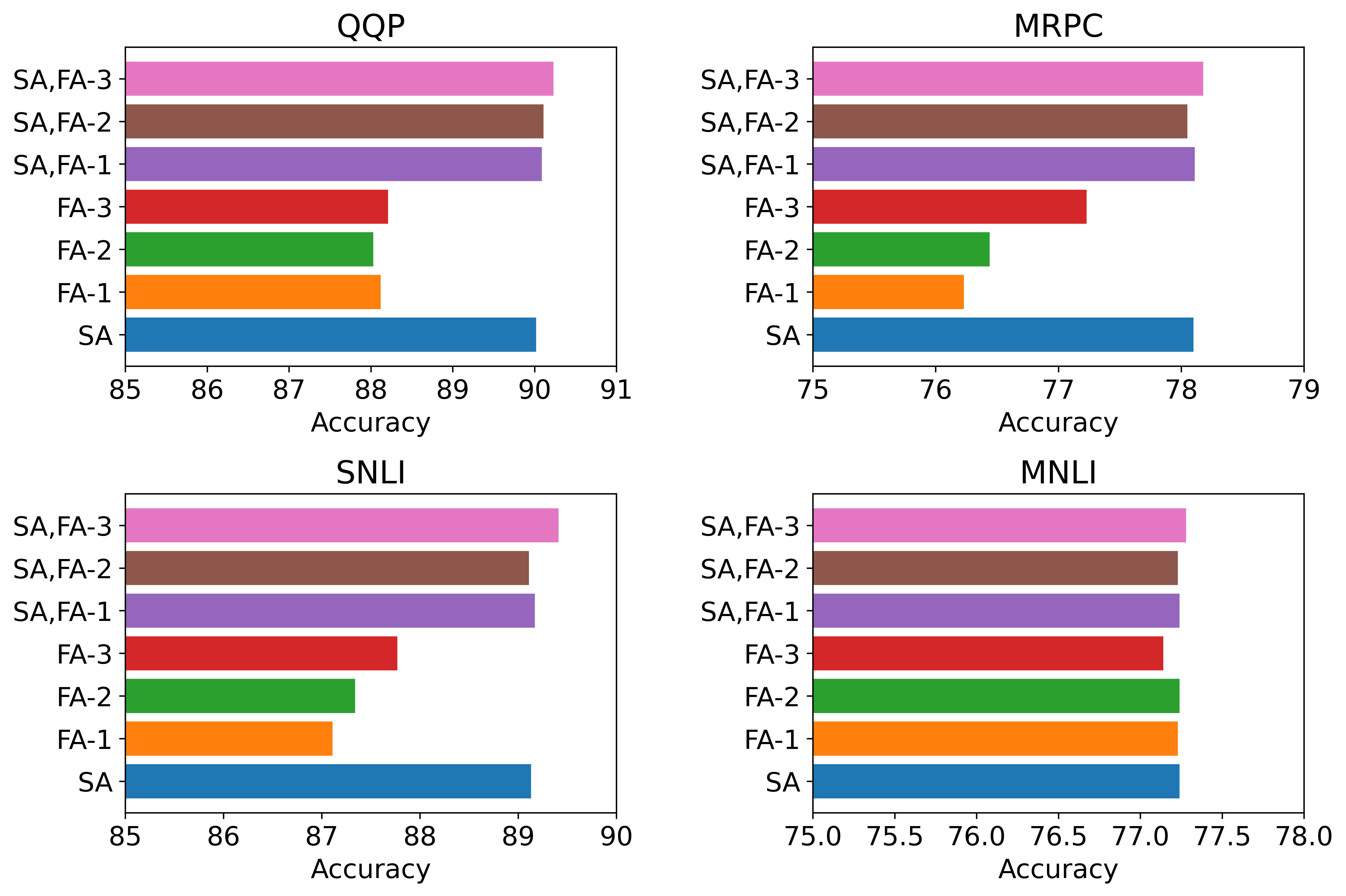}
\caption{Results of ablation study on various combination of strategy selections for SA and FA}
\label{fig.ab}
\end{figure}

\subsection{Implementation Details }
Our prediction target is ${\rm label^*}={\rm argmax}{P({\rm label}|\bm{s}^x,\bm{s}^y)}$. For the QQP and MRPC datasets, ${\rm label^*}$ represents \{match, not match\}, while for the SNLI and MNLI datasets, ${\rm label^*}$ corresponds to \{entailment, neutral, contradiction\}.
For all baselines, as well as our models, we use BERT (Base) as the encoder with 12 Transformer blocks, resulting in an encoded feature dimension of 768. Each sentence is padded to the average length of sentences in the dataset. To minimize the impact of the number of parameters on model performance, we adjusted the hyper-parameters of the convolutional and fully connected layers in the model, keeping the training parameters of all Siamese-based baselines and our model at similar levels. In the calculation of inference latency, the values we display are normalized by the test set size. We train the model for 15 epochs with a batch size of 32 and a learning rate of 0.001. Cross-entropy is used as the loss function, and the Adam optimizer with $\beta_1 = 0.9$ and $\beta_2 = 0.999$ is employed.  We adopt a learning rate decay strategy and early stopping here. Specifically, if there is no early stopping after 10 epochs, the learning rate will be reduced with a decay rate of 0.1.

\section{Results and Analysis}

\subsection{Main Results}

Table~\ref{tab.main} presents the main results for both baselines and our proposed approach across four evaluation tasks (QQP, MRPC, SNLI, MNLI). The principal difference between SBERT and ColBERT lies in the latter's implementation of Late Attention (LA), which contributes to ColBERT's enhanced performance. Our model is built upon various combination strategies of the aforementioned modules, utilizing proposed modules such as AFE for feature extraction, SA, FA-1, FA-2, and FA-3. In terms of feature fusion, our model explores both max pooling and average pooling, as well as RFM. The experimental results indicate that our model surpasses the baseline methods in terms of accuracy for the majority of tasks. Models employing RFM for feature fusion exhibit higher accuracy compared to those utilizing max pooling and average pooling. This can be attributed to the increased receptive field provided by RFM. Among the various attention mechanism combinations, the pairing of SA and FA-3 demonstrates the most favorable performance. BERT (interaction-based), as a powerful interactive-based model, outperforms our model in accuracy. However, its intricate interaction pattern leads to a substantially greater inference latency and parameter count relative to our method.

In conclusion, our network demonstrates superior performance in terms of accuracy across four benchmark tasks when compared to the SBERT and ColBERT. Compared with BERT (interaction-based), our network exhibits significant advantages in aspects such as inference latency and parameter count. Consequently, our network requires less storage space and computational resources, rendering it more valuable for real-time applications and low-latency scenarios.
The modular design of our network allows for the implementation of various combination strategies based on the selection of different components. These components possess minimal requirements for inter-module communication, facilitating seamless integration and enhancing the model's modularity and scalability. This adaptable design paves the way for further optimization and refinement of the model, ultimately contributing to advancements in the field.

\begin{figure}[t]
\centering
\includegraphics[width=1\linewidth]{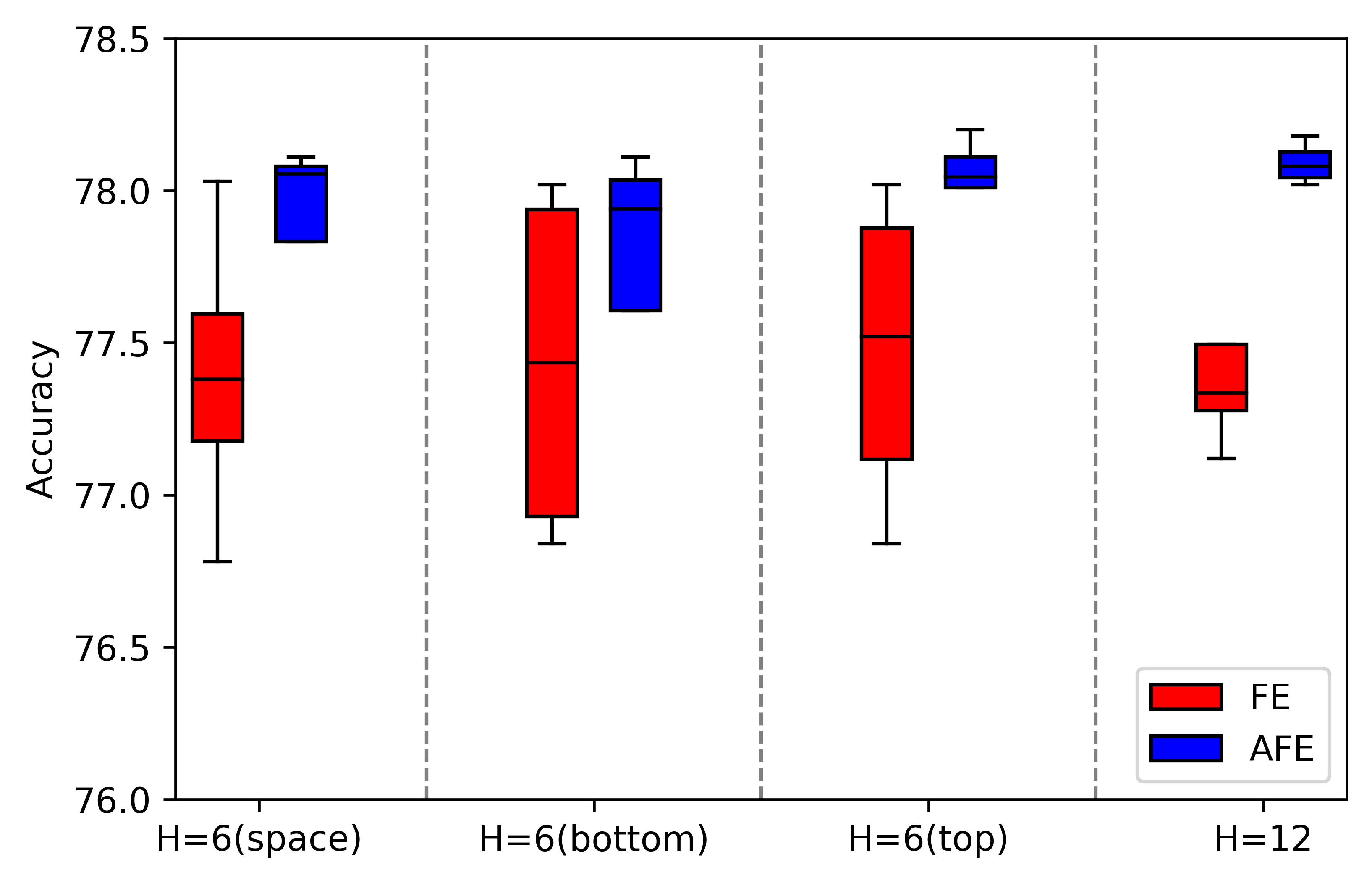}
\caption{Robust experimental results on the impact of Transformer block and adaptive weight selection for feature extraction. FE and AFE represent feature extraction without the introduction of adaptive weights and feature extraction with the incorporation of adaptive weights, respectively.}
\label{fig.robust}
\end{figure}

\subsection{Ablation Study}

\begin{figure*}[t]
\centering
\includegraphics[width=1\textwidth]{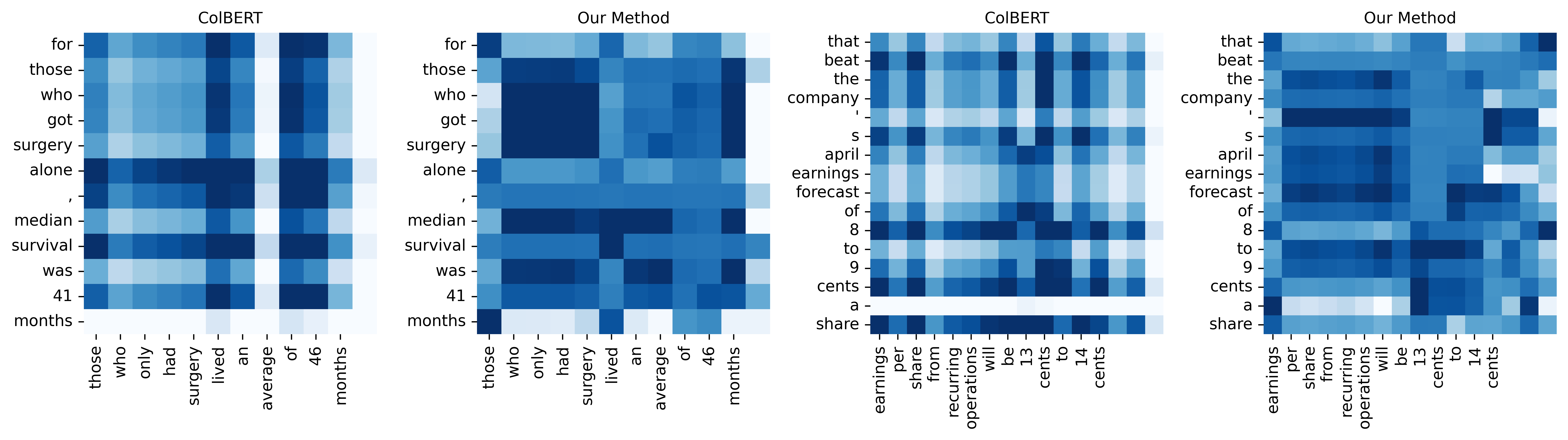}
\caption{Both ColBERT and our method encode the similarity matrix of sentence pairs. The darker the color in the image, the higher the degree of similarity.}\label{fig.case}
\end{figure*}

\subsubsection{Effect of SA\&FA}
Spatial Attention and Feature Attention together form a powerful information interactor. To gain a clear understanding of the impact of different attention combinations on model performance, we conducted ablation study on various combination strategy selections for SA and FA. As illustrated in Figure~\ref{fig.ab}, the standalone SA consistently outperforms the independent FA-1, FA-2, and FA-3 across all tasks. For QQP, MRPC, and SNLI tasks, combining SA with FA-1, FA-2, and FA-3 results in a significant performance boost. In the absence of SA, the differences in performance among FA-1, FA-2, and FA-3 are not substantial. Upon incorporating SA, the SA and FA-3 combination achieves the best results in all four tasks. In summary, SA serves as the foundation for improving the modeling of semantic similarity, and the SA, FA-3 approach demonstrates more stable and superior performance compared to other methods. This suggests that combining SA and FA-3 methodologies may lead to better extraction and utilization of feature information in these tasks.

\subsubsection{Effect of RFM}

The superior performance of the RFM is due to the introduction of Inception architecture and Dilated Convolutions to increase the network's receptive field. In this ablation study on QQP and MRPC datasets, we investigate the impact of Inception architecture and Dilated Convolutions on the network's capabilities. Specifically, for the Inception architecture, we only introduce multi-branch convolutions with different kernel sizes; for Dilated Convolutions, we only employ multi-branch Dilated Convolutions with different dilation rates. In the experiments, we use global max pooling and average pooling structures as references. As can be seen from Table~\ref{tab.ab}, both Inception and Dilated Convolution are able to enhance the network's performance.

\begin{table}
\begin{tabular}{l|l|l|l}
\hline
Architecture   & QQP                 & MRPC                 & Params              \\ \hline
Max\&Avg Pooling & 89.46               & 77.04               & 6.1M               \\ 
Inception        & 89.73($\uparrow$0.27) & 77.86($\uparrow$0.82) & 6.1M \\ 
Dilated Convolution     & 90.12($\uparrow$0.66) & 77.91($\uparrow$0.87) & 8.2M\\ 
RFM              & 90.23($\uparrow$0.77) & 78.18($\uparrow$1.14) & 6.4M \\ \hline
\end{tabular}
\caption{Ablation study on different architectures in RFM}~\label{tab.ab}
\end{table}

\subsection{Robustness Experiments}

Our method demonstrates exceptional performance, which can be attributed to the efficient and comprehensive extraction of information from Transformer blocks. In this experiment, we investigated the impact of varying the number of retained Transformer blocks during feature extraction and the incorporation of adaptive weights on model performance. As reported in Figure~\ref{fig.robust}, we conducted eight sets of experiments on MRPC dataset, examining the differences in experimental results after applying adaptive weights to four different model strategies with various Transformer block configurations: 6 spaced Transformer blocks (selecting Transformer blocks at intervals), 6 bottom Transformer blocks, 6 top Transformer blocks, and 12 Transformer blocks. When comparing different choices of Transformer blocks without introducing adaptive weights, we observed that selecting all Transformer blocks or retaining the top 6 Transformer blocks resulted in better model performance and stability than retaining the bottom 6 Transformer blocks or spacing 6 Transformer blocks. This is likely because the top Transformer blocks capture more comprehensive spatial and feature domain information after passing through multiple attention and feedforward layers. Introducing adaptive weights not only significantly enhanced the effectiveness of feature extraction but also minimized fluctuations between different network combination strategies. Furthermore, it reduced the influence of the number of Transformer blocks on the model, ultimately bolstering the network's robustness.

\subsection{Case Study}

To illustrate the advantages of our 3D semantic similarity modeling framework compared to traditional models, we encoded sentences from the MRPC dataset using both the pre-trained ColBERT model and our own model, obtaining semantic similarity matrices for sentence pairs. Two examples are displayed in the Figure~\ref{fig.case}, along with visualizations of their attention similarity matrices. ColBERT can solely focus on parts of the sentence pairs with similar meanings, such as 'survival', 'live', '8 to 9 cents', and '13 to 14 cents'. In contrast, our approach can pay more attention to key semantic information like 'who got surgery', 'who only had surgery', 'average', 'median', 'beat the company's April earnings cast', and 'earnings per share from recurring operations'. This demonstrates that our method, which efficiently utilizes raw information and models semantic similarity in a three-dimensional manner, can capture focus information more effectively.

\section{Conclusion}
In this paper, we propose a novel three-dimensional Siamese network for modeling semantic similarity. To reinforce this 3D framework, we have introduced a series of modules that address three key aspects: feature extraction, attention, and feature fusion. Extensive experiments on four text semantic similarity benchmarks demonstrate the effectiveness and efficiency of this 3D Siamese Network. Moreover, our introduced modules exhibit a "plug-and-play" characteristic, contributing to the model's robust modularity and scalability.

In the future, we plan to apply this concept of three-dimensional semantic modeling to other tasks within the field of natural language processing.

\bibliography{ecai}

\begin{thebibliography}{10}

\bibitem{bowman2015large}
Samuel Bowman, Gabor Angeli, Christopher Potts, and Christopher~D Manning, `A
  large annotated corpus for learning natural language inference', in {\em
  Proceedings of the 2015 Conference on Empirical Methods in Natural Language
  Processing}, pp. 632--642, (2015).

\bibitem{cao2020deformer}
Qingqing Cao, Harsh Trivedi, Aruna Balasubramanian, and Niranjan
  Balasubramanian, `Deformer: Decomposing pre-trained transformers for faster
  question answering', in {\em Proceedings of the 58th Annual Meeting of the
  Association for Computational Linguistics}, pp. 4487--4497, (2020).

\bibitem{cer2018universal}
Daniel Cer, Yinfei Yang, Sheng-yi Kong, Nan Hua, Nicole Limtiaco, Rhomni~St
  John, Noah Constant, Mario Guajardo-Cespedes, Steve Yuan, Chris Tar, et~al.,
  `Universal sentence encoder for english', in {\em Proceedings of the 2018
  Conference on Empirical Methods in Natural Language Processing: System
  Demonstrations}, pp. 169--174, (2018).

\bibitem{chen2017deeplab}
Liang-Chieh Chen, George Papandreou, Iasonas Kokkinos, Kevin Murphy, and Alan~L
  Yuille, `Deeplab: Semantic image segmentation with deep convolutional nets,
  atrous convolution, and fully connected crfs', {\em IEEE transactions on
  pattern analysis and machine intelligence}, {\bf 40}(4),  834--848, (2017).

\bibitem{chen2018encoder}
Liang-Chieh Chen, Yukun Zhu, George Papandreou, Florian Schroff, and Hartwig
  Adam, `Encoder-decoder with atrous separable convolution for semantic image
  segmentation', in {\em Proceedings of the European conference on computer
  vision (ECCV)}, pp. 801--818, (2018).

\bibitem{chen2017enhanced}
Qian Chen, Xiaodan Zhu, Zhen-Hua Ling, Si~Wei, Hui Jiang, and Diana Inkpen,
  `Enhanced lstm for natural language inference', in {\em Proceedings of the
  55th Annual Meeting of the Association for Computational Linguistics (Volume
  1: Long Papers)}, pp. 1657--1668, (2017).

\bibitem{devlin2019bert}
Jacob Devlin, Ming-Wei Chang, Kenton Lee, and Kristina Toutanova, `Bert:
  Pre-training of deep bidirectional transformers for language understanding',
  in {\em Proceedings of the 2019 Conference of the North American Chapter of
  the Association for Computational Linguistics: Human Language Technologies,
  Volume 1 (Long and Short Papers)}, pp. 4171--4186, (2019).

\bibitem{dolan2005automatically}
Bill Dolan and Chris Brockett, `Automatically constructing a corpus of
  sentential paraphrases', in {\em Third International Workshop on Paraphrasing
  (IWP2005)}. Asia Federation of Natural Language Processing, (January 2005).

\bibitem{gongnatural}
Yichen Gong, Heng Luo, and Jian Zhang, `Natural language inference over
  interaction space', in {\em International Conference on Learning
  Representations}.

\bibitem{gu2020context}
Zhangxuan Gu, Siyuan Zhou, Li~Niu, Zihan Zhao, and Liqing Zhang, `Context-aware
  feature generation for zero-shot semantic segmentation', in {\em Proceedings
  of the 28th ACM International Conference on Multimedia}, pp. 1921--1929,
  (2020).

\bibitem{heikkila2009description}
Marko Heikkil{\"a}, Matti Pietik{\"a}inen, and Cordelia Schmid, `Description of
  interest regions with local binary patterns', {\em Pattern recognition}, {\bf
  42}(3),  425--436, (2009).

\bibitem{hou2021coordinate}
Qibin Hou, Daquan Zhou, and Jiashi Feng, `Coordinate attention for efficient
  mobile network design', in {\em Proceedings of the IEEE/CVF Conference on
  Computer Vision and Pattern Recognition}, pp. 13713--13722, (2021).

\bibitem{hu2018squeeze}
Jie Hu, Li~Shen, and Gang Sun, `Squeeze-and-excitation networks', in {\em
  Proceedings of the IEEE conference on Computer Vision and Pattern
  Recognition}, pp. 7132--7141, (2018).

\bibitem{huang2017densely}
Gao Huang, Zhuang Liu, Laurens Van Der~Maaten, and Kilian~Q Weinberger,
  `Densely connected convolutional networks', in {\em Proceedings of the IEEE
  conference on computer vision and pattern recognition}, pp. 4700--4708,
  (2017).

\bibitem{huang2013learning}
Po-Sen Huang, Xiaodong He, Jianfeng Gao, Li~Deng, Alex Acero, and Larry Heck,
  `Learning deep structured semantic models for web search using clickthrough
  data', in {\em Proceedings of the 22nd ACM International Conference on
  Information \& Knowledge Management}, pp. 2333--2338, (2013).

\bibitem{humeau2019poly}
Samuel Humeau, Kurt Shuster, Marie-Anne Lachaux, and Jason Weston,
  `Poly-encoders: Architectures and pre-training strategies for fast and
  accurate multi-sentence scoring', in {\em Proceedings of the 8th
  International Conference on Learning Representations (ICLR 2020)}, (2020).

\bibitem{iyer2017first}
Shankar Iyer, Nikhil Dandekar, Korn{\'e}l Csernai, et~al., `First quora dataset
  release: Question pairs. data. quora. com', (2017).

\bibitem{khattab2020colbert}
Omar Khattab and Matei Zaharia, `Colbert: Efficient and effective passage
  search via contextualized late interaction over bert', in {\em Proceedings of
  the 43rd International ACM SIGIR conference on research and development in
  Information Retrieval}, pp. 39--48, (2020).

\bibitem{li2021virt}
Dan Li, Yang Yang, Hongyin Tang, Jingang Wang, Tong Xu, Wei Wu, and Enhong
  Chen, `{VIRT:} improving representation-based models for text matching
  through virtual interaction', {\em CoRR}, {\bf abs/2112.04195}, (2021).

\bibitem{liu2021pay}
Hanxiao Liu, Zihang Dai, David So, and Quoc~V Le, `Pay attention to mlps', {\em
  Advances in Neural Information Processing Systems}, {\bf 34},  9204--9215,
  (2021).

\bibitem{liu2018receptive}
Songtao Liu, Di~Huang, et~al., `Receptive field block net for accurate and fast
  object detection', in {\em Proceedings of the European conference on computer
  vision (ECCV)}, pp. 385--400, (2018).

\bibitem{ni2022sentence}
Jianmo Ni, Gustavo~Hernandez Abrego, Noah Constant, Ji~Ma, Keith Hall, Daniel
  Cer, and Yinfei Yang, `Sentence-t5: Scalable sentence encoders from
  pre-trained text-to-text models', in {\em Findings of the Association for
  Computational Linguistics: ACL 2022}, pp. 1864--1874, (2022).

\bibitem{nogueira2019passage}
Rodrigo~Frassetto Nogueira and Kyunghyun Cho, `Passage re-ranking with {BERT}',
  {\em CoRR}, {\bf abs/1901.04085}, (2019).

\bibitem{palangi2016deep}
Hamid Palangi, Li~Deng, Yelong Shen, Jianfeng Gao, Xiaodong He, Jianshu Chen,
  Xinying Song, and Rabab Ward, `Deep sentence embedding using long short-term
  memory networks: Analysis and application to information retrieval', {\em
  IEEE/ACM Transactions on Audio, Speech, and Language Processing}, {\bf
  24}(4),  694--707, (2016).

\bibitem{raffel2020exploring}
Colin Raffel, Noam Shazeer, Adam Roberts, Katherine Lee, Sharan Narang, Michael
  Matena, Yanqi Zhou, Wei Li, and Peter~J Liu, `Exploring the limits of
  transfer learning with a unified text-to-text transformer', {\em Journal of
  Machine Learning Research}, {\bf 21},  1--67, (2020).

\bibitem{reimers2019sentence}
Nils Reimers and Iryna Gurevych, `Sentence-bert: Sentence embeddings using
  siamese bert-networks', in {\em Proceedings of the 2019 Conference on
  Empirical Methods in Natural Language Processing and the 9th International
  Joint Conference on Natural Language Processing (EMNLP-IJCNLP)}, pp.
  3982--3992, (2019).

\bibitem{ruckle2020multicqa}
Andreas R{\"u}ckl{\'e}, Jonas Pfeiffer, and Iryna Gurevych, `Multicqa:
  Zero-shot transfer of self-supervised text matching models on a massive
  scale', in {\em Proceedings of the 2020 Conference on Empirical Methods in
  Natural Language Processing (EMNLP)}, pp. 2471--2486, (2020).

\bibitem{szegedy2015going}
Christian Szegedy, Wei Liu, Yangqing Jia, Pierre Sermanet, Scott Reed, Dragomir
  Anguelov, Dumitru Erhan, Vincent Vanhoucke, and Andrew Rabinovich, `Going
  deeper with convolutions', in {\em Proceedings of the IEEE Conference on
  Computer Vision and Pattern Recognition}, pp. 1--9, (2015).

\bibitem{szegedy2016rethinking}
Christian Szegedy, Vincent Vanhoucke, Sergey Ioffe, Jon Shlens, and Zbigniew
  Wojna, `Rethinking the inception architecture for computer vision', in {\em
  Proceedings of the IEEE Conference on Computer Vision and Pattern
  Recognition}, pp. 2818--2826, (2016).

\bibitem{vaswani2017attention}
Ashish Vaswani, Noam Shazeer, Niki Parmar, Jakob Uszkoreit, Llion Jones,
  Aidan~N Gomez, {\L}ukasz Kaiser, and Illia Polosukhin, `Attention is all you
  need', {\em Advances in Neural Information Processing Systems}, {\bf 30},
  (2017).

\bibitem{wang2017bilateral}
Zhiguo Wang, Wael Hamza, and Radu Florian, `Bilateral multi-perspective
  matching for natural language sentences', in {\em Proceedings of the 26th
  International Joint Conference on Artificial Intelligence}, pp. 4144--4150,
  (2017).

\bibitem{williams2018broad}
Adina Williams, Nikita Nangia, and Samuel~R Bowman, `A broad-coverage challenge
  corpus for sentence understanding through inference', in {\em 2018 Conference
  of the North American Chapter of the Association for Computational
  Linguistics: Human Language Technologies, NAACL HLT 2018}, pp. 1112--1122.
  Association for Computational Linguistics (ACL), (2018).

\bibitem{woo2018cbam}
Sanghyun Woo, Jongchan Park, Joon-Young Lee, and In~So Kweon, `Cbam:
  Convolutional block attention module', in {\em Proceedings of the European
  Conference on Computer Vision (ECCV)}, pp. 3--19, (2018).

\bibitem{xiong2020approximate}
Lee Xiong, Chenyan Xiong, Ye~Li, Kwok-Fung Tang, Jialin Liu, Paul~N Bennett,
  Junaid Ahmed, and Arnold Overwijk, `Approximate nearest neighbor negative
  contrastive learning for dense text retrieval', in {\em Proceedings of the
  8th International Conference on Learning Representations (ICLR 2020)},
  (2020).

\bibitem{yang2019simple}
Runqi Yang, Jianhai Zhang, Xing Gao, Feng Ji, and Haiqing Chen, `Simple and
  effective text matching with richer alignment features', in {\em Proceedings
  of the 57th Annual Meeting of the Association for Computational Linguistics},
  pp. 4699--4709, (2019).

\end{thebibliography}
\end{document}